\documentclass{article}
\usepackage{hyperref}
\usepackage{microtype}
\usepackage[T1]{fontenc}    %
\usepackage{xcolor}
\usepackage{url}
\usepackage{bm}
\usepackage{booktabs}
\usepackage{multicol}
\usepackage{mathtools}
\usepackage{minibox}
\usepackage{graphicx}
\usepackage{tikz}
\usepackage{tikz-dependency}
\usepackage{etoolbox}
\usepackage{subcaption}
\usepackage{xspace}

\usepackage[accepted]{icml2018}
\usepackage{inconsolata}

\usetikzlibrary{arrows.meta}

\definecolor{supgray}{gray}{0.4}
\colorlet{darkblue}{blue!50!black}
\definecolor{mylightred}{rgb}{0.89,0.10,0.11}
\definecolor{mylightblue}{rgb}{0.22,0.49,0.72}
\definecolor{mylightgreen}{rgb}{0.30,0.69,0.29}
\definecolor{mylightpurple}{rgb}{0.60,0.31,0.64}
\colorlet{myred}{mylightred!80!black}
\colorlet{myblue}{mylightblue!75!black}
\colorlet{mygreen}{mylightgreen!50!black}
\colorlet{mypurple}{mylightpurple!60!black}

\definecolor{tab20n0}{HTML}{5154A3}
\definecolor{tab20n1}{HTML}{5060A2}
\definecolor{tab20n2}{HTML}{5074A2}
\definecolor{tab20n3}{HTML}{5087A2}
\definecolor{tab20n4}{HTML}{5095A2}

\newcommand{\tolowercase}[1]
{
{\ifthenelse{\equal{#1}{\Pi}}{\pi}{\MakeLowercase{#1}}}
}

\makeatletter
\newcommand\RedeclareMathOperator{%
  \@ifstar{\def\rmo@s{m}\rmo@redeclare}{\def\rmo@s{o}\rmo@redeclare}%
}
\newcommand\rmo@redeclare[2]{%
  \begingroup \escapechar\m@ne\xdef\@gtempa{{\string#1}}\endgroup
  \expandafter\@ifundefined\@gtempa
     {\@latex@error{\noexpand#1undefined}\@ehc}%
     \relax
  \expandafter\rmo@declmathop\rmo@s{#1}{#2}}
\newcommand\rmo@declmathop[3]{%
  \DeclareRobustCommand{#2}{\qopname\newmcodes@#1{#3}}%
}
\@onlypreamble\RedeclareMathOperator
\makeatother

\RedeclareMathOperator*{\arg}{\mathsf{arg}}
\RedeclareMathOperator*{\max}{\mathsf{max}}
\RedeclareMathOperator*{\min}{\mathsf{min}}
\RedeclareMathOperator*{\ln}{\mathsf{ln}}
\RedeclareMathOperator*{\exp}{\mathsf{exp}}
\DeclareMathAlphabet{\mathbb}{U}{msb}{m}{n}
\DeclareMathOperator{\softmax}{\mathsf{softmax}}
\DeclareMathOperator{\sparsemax}{\mathsf{sparsemax}}
\DeclareMathOperator{\map}{\mathsf{MAP}}
\DeclareMathOperator{\Marginal}{\mathsf{Marginal}}
\DeclareMathOperator{\smapop}{\mathsf{SparseMAP}}
\DeclareMathOperator*{\argmin}{\mathsf{arg\,min}}
\DeclareMathOperator*{\argmax}{\mathsf{arg\,max}}
\newcommand\smap{\ensuremath{\smapop}\xspace}

\newcommand{\bs}[1]{\bm{#1}}
\newcommand{\norm}[1]{\left\lVert#1\right\rVert}
\newcommand{\Simplex}{\triangle}
\newcommand{\tr}{\top}
\newcommand{\secref}[1]{\S\ref{sec:#1}}
\newcommand{\pfrac}[2]{\frac{\partial #1}{\partial #2}}
\newcommand\col[2]{\bs{\tolowercase{#1}}_{#2}}

\newcommand{\supp}{\mathcal{I}}

\newcommand{\pt}{\bs{\theta}}  %
\newcommand{\pr}{\bs{\eta}}  %

\newcommand*{\eg}{\textit{e.g.}\@\xspace}
\newcommand*{\ie}{\textit{i.e.}\@\xspace}

\newtheorem{definition}{Definition} 
\newtheorem{proposition}{Proposition} 

\newenvironment{itemizesquish}{\begin{list}{\labelitemi}{\setlength{\topsep}{0em}\setlength{\itemsep}{0em}\setlength{\labelwidth}{0.75em}\setlength{\leftmargin}{\labelwidth}\addtolength{\leftmargin}{\labelsep}}}{\end{list}}

\newenvironment{enumeratesquish}{\setcounter{enumi}{0}\begin{list}{\addtocounter{enumi}{1}\labelenumi}{\setlength{\topsep}{0em}\setlength{\itemsep}{0em}\setlength{\labelwidth}{0.75em}\setlength{\leftmargin}{\labelwidth}\addtolength{\leftmargin}{\labelsep}}}{\end{list}\setcounter{enumi}{0}}

\icmltitlerunning{SparseMAP: Differentiable Sparse Structured Inference}

\begin{document}

\twocolumn[
\icmltitle{SparseMAP: Differentiable Sparse Structured Inference}

\begin{icmlauthorlist}
\icmlauthor{Vlad Niculae}{cor}
\icmlauthor{Andr\'e F.\,T. Martins}{ist}
\icmlauthor{Mathieu Blondel}{ntt}
\icmlauthor{Claire Cardie}{cor}
\end{icmlauthorlist}

\icmlaffiliation{cor}{Cornell University, Ithaca, NY}
\icmlaffiliation{ist}{Unbabel \& Instituto de Telecomunica\c{c}\~oes, Lisbon, Portugal}
\icmlaffiliation{ntt}{NTT Communication Science Laboratories, Kyoto, Japan}

\icmlcorrespondingauthor{Vlad Niculae}{\href{mailto:vlad@vene.ro}{\tt vlad@vene.ro}}
\icmlcorrespondingauthor{Andr\'e F.\,T. Martins}{\href{mailto:andre.martins@unbabel.com}{\tt andre.martins@unbabel.com}}
\icmlcorrespondingauthor{Mathieu Blondel}{\href{mailto:mathieu@mblondel.org}{\tt mathieu@mblondel.org}}
\icmlcorrespondingauthor{Claire Cardie}{\href{mailto:cardie@cs.cornell.edu}{\tt cardie@cs.cornell.edu}}

\icmlkeywords{Structured Prediction, Structured Inference, Sparsity, Machine Learning, ICML}

\vskip 0.3in
]

\printAffiliationsAndNotice{}

\begin{abstract}
Structured prediction requires searching over a combinatorial number of structures. 
To tackle it,
we introduce {\smap}: a new method for {\bf sparse structured inference},
and its natural loss function. 
{\smap}
automatically selects only a few global structures:
it is situated
between MAP inference, which picks a single structure,
and marginal inference, which assigns nonzero probability to all structures, 
including implausible ones. 
{\smap} can be computed using only calls to a MAP oracle, 
making it applicable to problems with intractable marginal inference, \eg,
linear assignment. 
Sparsity makes gradient backpropagation efficient
regardless of the structure, 
enabling us to augment deep neural networks with generic and sparse
{\bf structured hidden layers}.
Experiments in dependency parsing 
and natural language inference 
reveal competitive accuracy, improved interpretability, 
and the ability to capture 
natural language ambiguities,
which is attractive for pipeline systems.
\end{abstract}

\section{Introduction}

{\bf Structured prediction} involves the manipulation of discrete,
combinatorial structures, \eg, trees and alignments 
\citep{Bakir2007,Smith2011,Nowozin2014}. 
Such structures arise naturally 
as machine learning outputs,
and as intermediate
representations in deep pipelines. 
However, the set of possible structures is typically prohibitively large. As such,
inference is a core challenge, often sidestepped by greedy
search, factorization assumptions, or continuous relaxations
\citep{Belanger2016}. 

In this paper, we propose an
appealing alternative: a new inference strategy, dubbed {\bf \smap}, which encourages {\bf sparsity} in the structured representations. Namely, we
seek solutions explicitly expressed as a combination of a small, enumerable
set of global structures. 
\begin{figure}[t]
\centering
\hbox{%
\begin{tikzpicture}[every label/.style={align=center}]
   \small
   \newdimen\R
   \R=1.3cm
   \coordinate (a) at (0, 0);
   \coordinate (b) at (2*\R, 0);
   \coordinate (c) at (60:2*\R);
   \draw (a) -- (b) -- (c) -- cycle;
   \fill [opacity=0.03,orange] (a) -- (b) -- (c) -- cycle;

   \node[inner sep=1.2pt,circle,draw,fill,label={left:$\Simplex$~}] at (a) {};
   \node[inner sep=1.2pt,circle,draw,fill,label={}] at (b) {};
   \node[inner sep=2.2pt,circle,draw=gray,fill=myblue,label=right:{%
   \color{myblue} argmax\\{\scriptsize \color{myblue} (1, 0, 0)}%
   }] at (c) {};
   \node[inner
   sep=2.2pt,circle,draw=gray,fill=mypurple,label=below:{%
   \color{mypurple}softmax\\{\scriptsize \color{mypurple} (.5, .3, .2)}%
   }] at
   (45:1.3*\R) {};
   \node[inner sep=2.2pt,circle,draw=gray,fill=mygreen,label={[label
   distance=0.001cm]185:{%
   \color{mygreen}sparsemax\\{\scriptsize \color{mygreen} (.6, .4, 0)}
   }}]
   at (60:1.2*\R) {};
\end{tikzpicture}
\hspace{-0.3cm}%
\newsavebox{\treemap}
\savebox{\treemap}{%
\begin{tikzpicture}[scale=0.22,color=myblue]

\node[inner sep=1pt] (root) at (1,4) {$\star$};
\begin{scope}[every node/.style={inner sep=1pt,circle,draw}]
    \node (a) at (0,2) {};
    \node (b) at (2,2) {};
    \node (c) at (0,0) {};
\end{scope}

\begin{scope}[>={Stealth[myblue,length=1.3mm]},
              every edge/.style={draw}]
    \path [<-] (a) edge (root);
    \path [<-] (b) edge (root);
    \path [<-] (c) edge (a);
\end{scope}
\end{tikzpicture}
}

\newsavebox{\treesparsemap}
\savebox{\treesparsemap}{%
\begin{tikzpicture}[scale=0.22,color=mygreen]

\node[inner sep=1pt] (root) at (1,4) {$\star$};
\begin{scope}[every node/.style={inner sep=1pt,circle,draw}]
    \node (a) at (0,2) {};
    \node (b) at (2,2) {};
    \node (c) at (0,0) {};
\end{scope}

\begin{scope}[>={Stealth[mygreen,length=1.3mm]},
              every edge/.style={draw}]
    \path [<-] (a) edge (root);
    \path [<-] (b) edge (root);

\begin{scope}[transparency group, opacity=0.5]
    \path [<-] (c) edge (a);
    \path [<-] (c) edge (b);
\end{scope}

\end{scope}
\end{tikzpicture}
}

\newsavebox{\treemarginal}
\savebox{\treemarginal}{%
\begin{tikzpicture}[scale=0.22,color=mypurple]

\node[inner sep=1pt] (root) at (1,4) {$\star$};
\begin{scope}[every node/.style={inner sep=1pt,circle,draw}]
    \node (a) at (0,2) {};
    \node (b) at (2,2) {};
    \node (c) at (0,0) {};
\end{scope}

\begin{scope}[>={Stealth[mypurple,length=1.3mm]},
              every edge/.style={draw}]
\begin{scope}[transparency group, opacity=0.6]
    \path [<-] (a) edge (root);
    \path [<-] (b) edge (root);
\end{scope}
\begin{scope}[transparency group, opacity=0.25]
    \path [<-] (c) edge[bend left=75] (root);
\end{scope}

\begin{scope}[transparency group, opacity=0.5]
    \path [<->] (c) edge (a);
    \path [<->] (c) edge (b);
\end{scope}

\begin{scope}[transparency group, opacity=0.33]
    \path [<->] (b) edge (a);
\end{scope}

\end{scope}
\end{tikzpicture}
}
 \begin{tikzpicture}[every label/.style={align=center}]
   \small
   \newdimen\R
   \R=1.3cm
   \fill [opacity=0.03,orange]
       (0:\R) \foreach \x in {60,120,...,360} {  -- (\x:\R) };

   \draw (0:\R) \foreach \x in {60,120,...,360} {  -- (\x:\R) };

   \foreach \x/\l/\p in
     { 
      120/{}/above,
      180/{}/left,
      240/{$\mathcal{M}$~}/left,
      300/{}/below,
      360/{}/right
     }
     \node[inner sep=1.2pt,circle,draw,fill,label={\p:\l}] at (\x:\R) {};

     \node[inner sep=2.2pt,circle,draw=gray,fill=myblue,%
           label={[myblue,shift={(0.9cm,-0.3cm)}]MAP~\usebox{\treemap}}] at (60:\R) {};

     \node[inner sep=2.2pt,circle,draw=gray,fill=mypurple,%
           label={[mypurple,shift={(-0.4cm,-1.7cm)}]Marginal\\\usebox{\treemarginal}}]%
     at (60:0.5*\R) {};

     \node[inner sep=2.2pt,circle,draw=gray,fill=mygreen,,%
         label={[mygreen,shift={(1cm,-1.4cm)}]{{\smap}\\\usebox{\treesparsemap}}}]%
     at (38:0.87\R) {};
\end{tikzpicture}
}
\caption{\label{fig:sketch}%
Left: in the unstructured case, $\softmax$ and $\sparsemax$ can be interpreted
as regularized, differentiable $\argmax$ approximations; 
$\softmax$ returns dense solutions while $\sparsemax$ favors sparse ones.
Right: in this work, we extend this view to {\em structured inference}, which
consists of
optimizing over a polytope $\mathcal{M}$, the convex hull of
all possible structures (depicted: the arborescence polytope, whose vertices are
trees).  We introduce {\smap} as a structured extension of $\sparsemax$: it
is situated
in between MAP inference, which yields a single structure, and marginal
inference, which returns a dense combination of structures.
} 
\end{figure}
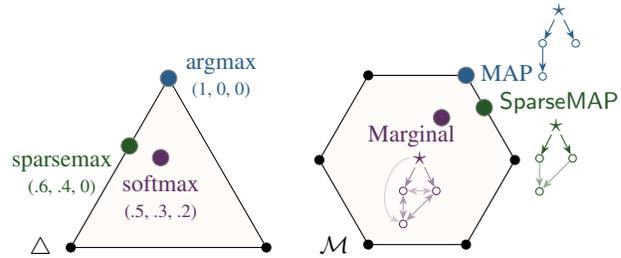
Our framework departs from the two most common inference strategies in structured prediction: {\bf
maximum {\em a posteriori} (MAP) inference}, which returns the highest-scoring
structure, and {\bf marginal inference}, which yields
a dense probability distribution over structures. 
Neither of these strategies is fully satisfactory: 
for latent structure models, marginal
inference is appealing, since it can represent uncertainty and, unlike MAP inference, it is continuous and differentiable, hence 
amenable for use in structured hidden layers in neural networks \citep{rush}. 
It has, however, several limitations. First, there are useful
problems for which MAP is tractable, but marginal inference is not, \eg, 
linear assignment
\citep{valiant1979complexity,taskar-thesis}.  Even when marginal inference
is available, case-by-case derivation of the backward pass is needed,
sometimes producing  fairly complicated algorithms, \eg,
second-order expectation semirings \citep{Li2009}.
Finally, marginal inference is {\em dense}:
it assigns nonzero probabilities to all structures and cannot {\em completely rule out}
irrelevant ones.  This can be statistically and computationally wasteful, as
well as qualitatively harder to interpret.

In this work, we make the following contributions:

\begin{enumeratesquish}
\item We propose {\smap}: a new framework for {\bf sparse structured inference}
(\secref{smap}).
The main idea is illustrated in Figure~\ref{fig:sketch}.
{\smap} is a twofold generalization:
first,
as a structured extension of
the $\sparsemax$ transformation \citep{sparsemax};
second,
as
a continuous yet sparse relaxation of MAP inference.
MAP yields a single structure and 
marginal inference yields a dense distribution over {\em all} structures. 
In contrast, the
{\smap} solutions are sparse combinations of a small number of often-overlapping structures.

\item We show how to {\bf compute {\smap} effectively},
requiring only a MAP solver as a subroutine (\secref{solve}), by exploiting the 
problem's sparsity and quadratic curvature.
Noticeably, 
the MAP oracle can be any arbitrary solver, \eg, the
Hungarian algorithm for linear assignment, 
which permits tackling problems for which marginal inference is intractable.

\item We derive expressions for {\bf gradient backpropagation} through {\smap}
inference, which, unlike MAP, is differentiable almost everywhere
(\secref{backpass}). 
The backward pass is fully general (applicable to any
type of structure), and it is efficient, thanks to the sparsity of the
solutions and to reusing quantities computed in the forward pass. 

\item We introduce a novel {\bf {\smap} loss} for structured prediction, 
placing it into a family of loss functions 
which generalizes the CRF and structured SVM losses (\secref{loss}). Inheriting
the desirable properties of {\smap} inference, the {\smap} loss and its
gradients can be computed efficiently, provided access to MAP inference.
\end{enumeratesquish}

Our experiments 
demonstrate that {\smap} is useful both for predicting structured outputs, as
well as for learning latent structured representations. 
On {\em dependency parsing} (\secref{parse}), structured output
networks trained with the {\smap} loss yield more accurate models
with sparse, interpretable predictions, adapting to the ambiguity
(or lack thereof) of test examples.
On {\em natural language inference} (\secref{esim}),
we learn latent structured alignments,
obtaining good predictive performance, as well as useful natural
visualizations concentrated on a small number of structures.%
\footnote{
General-purpose {\tt dynet} and {\tt pytorch} implementations
available at \url{https://github.com/vene/sparsemap}.}

\paragraph{Notation.}
Given vectors $\bs{a} \in \mathbb{R}^m, \bs{b} \in \mathbb{R}^n$,
$[\bs{a}; \bs{b}] \in \mathbb{R}^{m+n}$ denotes their concatenation;
given matrices $\bs{A} \in \mathbb{R}^{m \times k}, \bs{B} \in
\mathbb{R}^{n \times k}$, we denote their row-wise stacking as $[\bs{A}; \bs{B}]
\in \mathbb{R}^{(m+n) \times k}$. We denote the columns of a matrix $\bs{A}$ by
$\col{A}{j}$; by extension, a slice of columns of $\bs{A}$ is denoted
$\bs{A}_{\mathcal{I}}$ for a set of indices $\mathcal{I}$.
We denote
the canonical simplex by $\Simplex^d \coloneqq \{ \bs{y} \in \mathbb{R}^d \colon
\bs{y} \succeq \bs{0}, \sum_{i=1}^d y_i = 1\}$, and
the indicator function of a predicate $p$ as
$\mathbb{I}[p] = \{ 1 \text{ if } p,~0$ otherwise $\}$.

\section{Preliminaries}
\label{sec:prelim}

\subsection{Regularized Max Operators: Softmax, Sparsemax}
\label{sec:regmax}
As a basis for the more complex structured case, we first consider the simple problem of
selecting the largest value in a vector $\pt \in \mathbb{R}^d$.
We denote the vector mapping
\[
    \argmax(\pt) \coloneqq \argmax_{y \in \Simplex^d} \pt^\tr\bs{y}.
\]
When there are no ties, $\argmax$ has a unique solution $\bs{e}_i$ peaking at
the index $i$ of the highest value of $\pt$. When there are ties, $\argmax$ is
set-valued.  Even assuming no ties, $\argmax$ is piecewise constant, and thus
is ill-suited for direct use within neural networks, \eg, in an attention
mechanism. Instead, it is common to use $\softmax$, a continuous and
differentiable approximation to $\argmax$, which can be seen as an
entropy-regularized $\argmax$
\begin{equation}
    \softmax(\pt)
    \coloneqq
    \argmax_{\bs{y}\in\Simplex^d} \pt^\tr \bs{y} + H(\bs{y}) 
    =\frac{\exp \pt}{\sum_{i=1}^d \exp \theta_i} 
    \label{eqn:softmax} 
\end{equation}
where $H(\bs{y}) = -\sum_i y_i \ln y_i$,
\ie the negative Shannon entropy.
Since $\exp \cdot >0$ strictly, $\softmax$ outputs are dense.

By replacing the entropic penalty with a squared $\ell_2$ norm, \citet{sparsemax}
introduced a sparse alternative to $\softmax$, called $\sparsemax$, given by
\begin{equation}
\begin{aligned}
    \sparsemax(\pt) 
    \coloneqq& 
    \argmax_{\bs{y}\in\Simplex^d} \pt^\tr \bs{y} -\frac{1}{2}\norm{\bs{y}}_2^2 
    \\
    =& \argmin_{\bs{y} \in \Simplex^d} \norm{\bs{y} - \pt}_2^2. 
    \label{eqn:sparsemax} \\
\end{aligned}
\end{equation}
Both $\softmax$ and $\sparsemax$ are continuous and differentiable almost
everywhere; however, $\sparsemax$ encourages sparsity in its outputs. 
This is because it corresponds to an Euclidean projection onto the simplex, which is likely to hit its boundary as the magnitude of $\pt$ increases. 
Both mechanisms, as well as variants with different penalties~\cite{sparseattn},
have been successfully used in attention mechanisms, for mapping a score
vector $\pt$ to a $d$-dimensional normalized discrete probability distribution
over a small set of choices.
The relationship between $\argmax$, $\softmax$, and $\sparsemax$, illustrated
in Figure~\ref{fig:sketch}, sits at the foundation of \smap.

\subsection{Structured Inference}
\label{sec:structinf}
In structured prediction, the space of possible outputs is typically very large:
for instance, all possible labelings of a length-$n$ sequence, spanning
trees over $n$ nodes, or one-to-one alignments between two sets. 
We may still write optimization problems such as $\max_{s=1}^D \theta_s$,
but it is impractical to enumerate all of the $D$ possible structures
and, in turn, to specify the scores for each structure in $\pt$. 

Instead, structured problems are often parametrized through
{\bf structured log-potentials} (scores)
$\pt \coloneqq \bs{A}^\tr \pr$, where $\bs{A} \in \mathbb{R}^{k \times D}$ is a matrix that specifies the structure of the problem, and
$\pr \in \mathbb{R}^k$ is lower-dimensional parameter vector, \ie, $k \ll D$. 
For example, in a 
{\bf factor graph} 
\cite{fg} 
with variables $U$ and factors $F$,
$\pt$ is given by
\[
\theta_s \coloneqq \sum_{i\in U} \eta_{U, i}(s_i) + \sum_{f\in F} \eta_{F, f}(s_f),
\]
where $\bs{\eta}_U$ and $\bs{\eta}_F$ are unary and higher-order
log-potentials, and 
$s_i$ and $s_f$ are local configurations at variable and factor nodes. 
This can be written in matrix notation 
as $\pt = \bs{M}^\tr \bs{\eta}_U + \bs{N}^\tr \bs{\eta}_F$ 
for suitable matrices $\{\bs{M}, \bs{N}\}$, 
fitting the assumption above with $\bs{A}=[\bs{M};\bs{N}]$ and $\bs{\pr}=[\pr_U;\pr_F]$.

We can then  
rewrite the {\bf MAP inference} problem, 
which seeks the highest-scoring structure,  
as a $k$-dimensional problem, by introducing variables $[\bs{u};\bs{v}] \in \mathbb{R}^k$ to denote 
configurations at variable and factor nodes:%
\footnote{We use the notation $\argmax_{\bs{u}:~[\bs{u};\bs{v}] \in \mathcal{M}}$
to convey that the maximization is over both $\bs{u}$ and $\bs{v}$,
but only $\bs{u}$ is returned. 
Separating the variables as $[\bs{u};\bs{v}]$ loses no generality
and allows us to isolate the unary posteriors $\bs{u}$ as the return value of
interest.} %
\begin{equation}
\label{eqn:map}
    \begin{aligned}
    \map_{\bs{A}}(\pr)
    \coloneqq&
\argmax_{
\substack{\bs{u}\coloneqq\bs{My} \\ \bs{y}\in\Simplex^D
\\
\hphantom{\bs{u}:~[\bs{u};\bs{v}] \in \mathcal{M}_{\bs{A}}}
}
} 
    \pt^\tr\bs{y} \\
    =&
    \argmax_{\bs{u}:~[\bs{u};\bs{v}] \in \mathcal{M}_{\bs{A}}}
    \pr_U^\tr\bs{u} + \pr_F^\tr \bs{v},
\end{aligned}
\end{equation}
where $\mathcal{M}_{\bs{A}} \coloneqq \{
        [\bs{u}; \bs{v}]:
        \bs{u}=\bs{My},~
        \bs{v}=\bs{Ny},~
        \bs{y}\in\Simplex^D 
     \}$ is the {\bf marginal polytope} \citep{Wainwright2008},
with one vertex for each possible structure (Figure~\ref{fig:sketch}).
However, as previously said,
since it is equivalent to a $D$-dimensional $\argmax$, MAP is piecewise
constant and discontinuous. 

Negative entropy regularization over $\bs{y}$,
on the other hand, 
yields {\bf marginal inference},
\begin{equation}
\label{eqn:marginal}
    \begin{aligned}
    \Marginal_{\bs{A}}(\pr)
    \coloneqq& %
\argmax_{
\substack{\bs{u}\coloneqq\bs{My} \\ \bs{y}\in\Simplex^D
\\
\hphantom{\bs{u}:~[\bs{u};\bs{v}] \in \mathcal{M}_{\bs{A}}}
}
}
\pt^\tr\bs{y} + H(\bs{y}) \\
  =& \argmax_{\bs{u}:~[\bs{u};\bs{v}] \in \mathcal{M}_{\bs{A}}}
    \pr_U^\tr\bs{u} + \pr_F^\tr \bs{v}
    + H_{\bs{A}}(\bs{u}, \bs{v}).
\end{aligned}
\end{equation}
Marginal inference is differentiable, but may be more difficult to
compute;
the entropy $H_{\bs{A}}(\bs{u}, \bs{v})=H(\bs{y})$
itself lacks a closed form
\citep[\S 4.1.2]{Wainwright2008}.
Gradient backpropagation is available only to 
specialized problem instances, \eg those solvable by dynamic programming \citep{Li2009}. 
The entropic term regularizes $\bs{y}$ toward more uniform distributions,
resulting in strictly dense solutions, just like in the case of
$\softmax$~(Equation~\ref{eqn:softmax}).

Interesting types of structures, which we use in the experiments described in Section~\ref{sec:exp}, include
the following.

\textbf{Sequence tagging.}
    Consider a sequence of $n$ items, each assigned one out of a possible $m$ tags. In this case, a global structure $s$ is a joint assignment of tags $(t_1, \cdots, t_n)$.
    The matrix $\bs{M}$ is $nm$-by-$m^n$--dimensional, with columns
    $\col{M}{s} \in \{0, 1\}^{nm} \coloneqq [\bs{e}_{t_1}, ..., \bs{e}_{t_n}]$ indicating
    which tag is assigned to each variable in the global structure $s$.
    $\bs{N}$ is $nm^2$-by-$m^n$--dimensional,
    with $\col{N}{s}$ encoding the transitions between consecutive tags,
    \ie, $\col{N}{s}(i, a, b) \coloneqq \mathbb{I}[t_{i-1}=a~\&~t_{i}=b]$.
    The Viterbi algorithm provides MAP inference and forward-backward
    provides marginal inference \citep{Rabiner1989}.

\textbf{Non-projective dependency parsing.}
    Consider a sentence of length $n$. Here, a structure $s$
    is a dependency tree: a rooted spanning tree over the $n^2$ possible
    arcs (for example, the arcs above the sentences in
    Figure~\ref{fig:ambiparse}).  Each column $\col{M}{s} \in \{0,1\}^{n^2}$
    encodes a tree by assigning a $1$ to its arcs. $\bs{N}$ is
    empty, $\mathcal{M}_{\bs{A}}$ is known as the {\em arborescence polytope} 
    \cite{andre-concise}. MAP inference may be performed by
    {\em maximal arborescence} algorithms
    \cite{Chu1965,Edmonds1967,mcdonald2005online},
    and the Matrix-Tree theorem \citep{Kirchhoff1847} provides a way to perform marginal inference
    \cite{koo-mt,DSmithSmith2007}. %

\textbf{Linear assignment.} 
    Consider a one-to-one matching (linear assignment) between two sets
    of $n$ nodes. 
    A global structure $s$ is a $n$-permutation, and a column
    $\col{M}{s} \in \{0,1\}^{n^2}$ can be seen as a flattening of
    the corresponding permutation matrix. Again, $\bs{N}$ is empty.
    $\mathcal{M}_{\bs{A}}$ is the Birkhoff polytope
    \cite{birkhoff}, 
    and MAP inference can be performed by, \eg, the Hungarian
    algorithm~\cite{hungarian} or the Jonker-Volgenant
    algorithm~\cite{lapjv}.
    Noticeably, marginal inference is known to be \#P-complete
    \citep[Section 3.5]{valiant1979complexity,taskar-thesis}. This makes it an open problem how to use matchings as latent variables.

\section{{\boldmath \smap}}

Armed with the parallel between structured inference and regularized $\max$
operators described in \secref{prelim}, we are now ready to introduce {\smap},
a novel inference optimization problem which returns sparse solutions.
\subsection{Definition}
\label{sec:smap}
We introduce {\smap} by regularizing the MAP inference problem in
Equation~\ref{eqn:map} with a squared $\ell_2$ penalty on the
returned posteriors, \ie, $\frac{1}{2}\norm{\bs{u}}^2_2$. 
Denoting, as above, $\pt \coloneqq \bs{A}^\tr\pr$,
the result is a quadratic optimization problem,
\begin{equation}
\label{eqn:smap}
\begin{aligned}
    \smapop_{\bs{A}}(\pr) \coloneqq& 
    \argmax_{
\substack{\bs{u}\coloneqq\bs{My} \\ \bs{y}\in\Simplex^D \\
\hphantom{\bs{u}:~[\bs{u};\bs{v}] \in \mathcal{M}_{\bs{A}}}
}
} \pt^\tr \bs{y}
    - \frac{1}{2} \norm{\bs{My}}_2^2 \\
    =&
    \hspace{-0.15cm}
    \argmax_{\bs{u}:~[\bs{u}, \bs{v}] \in \mathcal{M}_{\bs{A}}}
    \pr_U^\tr\bs{u} + \pr_F^\tr \bs{v}
    - \frac{1}{2} \norm{\bs{u}}_2^2. \\
\end{aligned}
\end{equation}

The quadratic penalty replaces the entropic penalty from marginal inference
(Equation~\ref{eqn:marginal}), which pushes the solutions to the strict
interior of the marginal polytope. In consequence, $\smap$ favors sparse
solutions from the faces of the marginal polytope $\mathcal{M}_{\bs{A}}$,
as illustrated in Figure \ref{fig:sketch}.
For the structured prediction problems mentioned in Section~\ref{sec:structinf}, 
{\smap} would be able to return, for example, 
a sparse combination of sequence labelings, parse trees, or matchings. 
Moreover, the
strongly convex regularization on $\bs{u}$ ensures that {\smap} has a unique
solution and is differentiable almost everywhere, as we will see. 

\subsection{Solving {\boldmath \smap}}
\label{sec:solve}
\begin{figure}[t]
    \centering
    \includegraphics[width=0.485\textwidth]{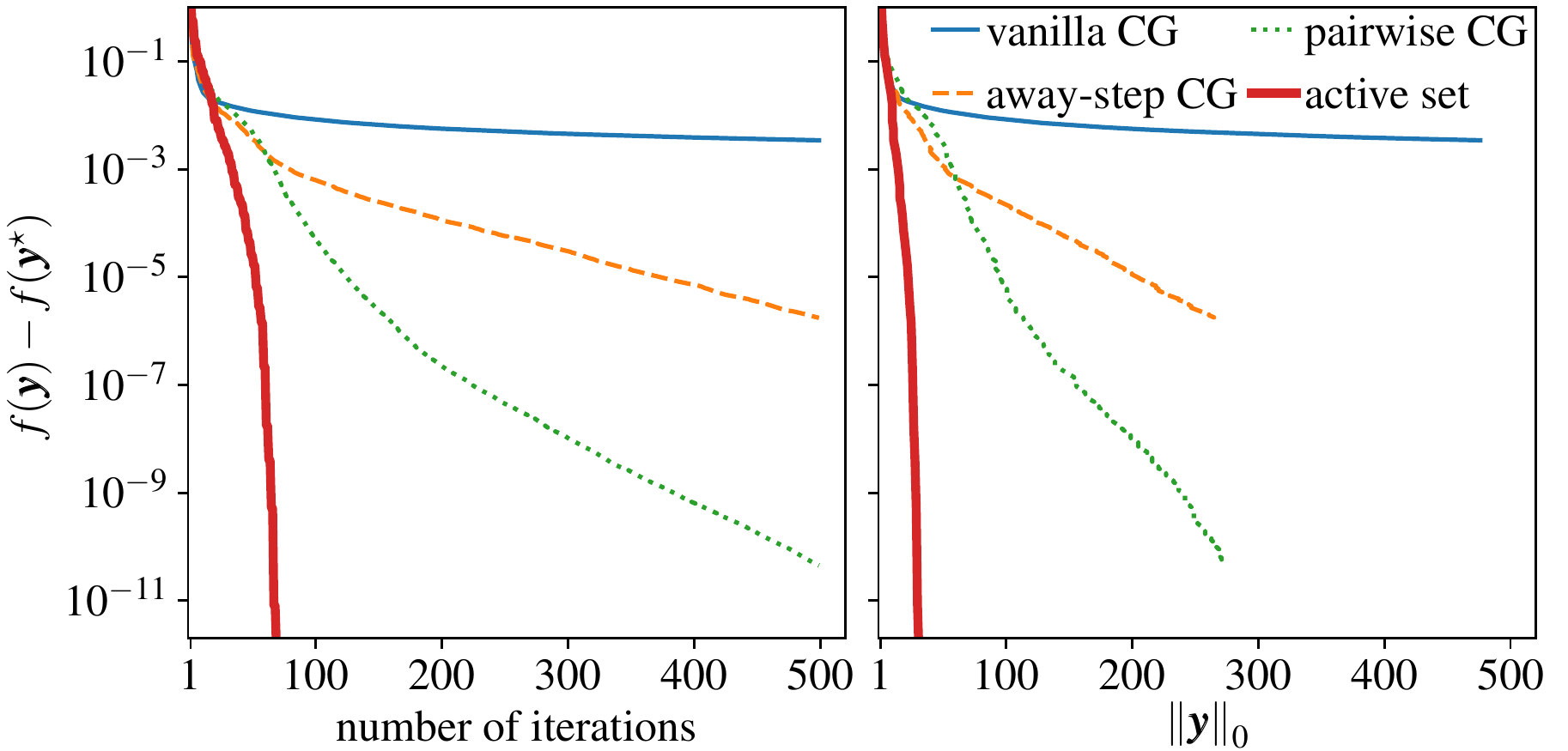}
    \caption{\label{fig:solver}Comparison of solvers on the {\smap}
    optimization problem for a tree factor with 20 nodes. The active
    set solver converges much faster and to a much sparser solution.}
\end{figure}
We now tackle the optimization problem in Equation~\ref{eqn:smap}. Although
{\smap} is a QP over a polytope, even describing it in standard form is
infeasible, since enumerating the exponentially-large set of vertices is
infeasible.  %
This prevents direct application of, \eg, the generic differentiable QP solver
of \citet{optnet}.
We instead focus on {\smap} solvers that involve a sequence of MAP problems as
a subroutine---this makes {\smap} widely applicable, given the availability of
MAP implementations for various structures.  We discuss two such methods, one
based on the conditional gradient algorithm and another based on the active set
method for quadratic programming.  We provide a full description of both
methods in Appendix~\ref{supp:solvers}.

\textbf{Conditional gradient.~} 
One family of such solvers is based on the {\em conditional gradient} (CG)
algorithm~\cite{fw,cg}, considered in prior work for solving approximations of
the marginal inference problem~\cite{belanger2013marginal,barrierfw}. 
Each step must solve a linearized subproblem. Denote by $f$ the {\smap}
objective from Equation~\ref{eqn:smap},
\[
    f(\bs{u}, \bs{v}) \coloneqq
    \pr_U^\tr\bs{u} + \pr_F^\tr \bs{v}
    - \frac{1}{2} \norm{\bs{u}}_2^2.
\]
The gradients of $f$ with respect to the two variables are
\[
    \nabla_{\bs{u}} f(\bs{u}', \bs{v}') = \pr_U - \bs{u}', \qquad 
    \nabla_{\bs{v}} f(\bs{u}', \bs{v}') = \pr_V.
\]
A linear approximation to $f$ around a point $[\bs{u}'; \bs{v}']$ is
\[
    \hat{f}(\bs{u}, \bs{v})\coloneqq 
    (\nabla_{\bs{u}}f)^\tr \bs{u}
    + (\nabla_{\bs{v}}f) ^\tr \bs{v}
    = (\pr_U - \bs{u}')^\tr \bs{u} + \pr_F^\tr \bs{v}.
\]
Minimizing $\hat{f}$ over $\mathcal{M}$ is exactly MAP inference with adjusted
variable scores $\pr_U - \bs{u}'$.
Intuitively, at each step we seek a high-scoring structure while penalizing
sharing variables with already-selected structures
Vanilla CG simply adds the new structure to the active set at every iteration.
The pairwise and away-step variants trade off between the
direction toward the new structure, and away from one of the already-selected
structures. More sophisticated variants have been proposed \citep{Garber2016}
which can provide sparse solutions when optimizing over a polytope. 

\textbf{Active set method.~}
Importantly, the {\smap} problem in Equation~\ref{eqn:smap} has quadratic
curvature, which the general CG algorithms may not optimally leverage. For this
reason, we consider the active set method for constrained QPs:
a generalization of Wolfe's min-norm point algorithm \citep{mnp},
also used in structured prediction for the quadratic subproblems by \citet{ad3}.
The active set algorithm, at
each iteration, updates an estimate of the solution support by adding or
removing one constraint to/from the active set; then it solves the
Karush--Kuhn--Tucker (KKT) system of a relaxed QP restricted to the current
support.  %

\textbf{Comparison.}
Both algorithms enjoy global linear convergence with similar rates
\citep{cg}, but the active set algorithm also exhibits exact finite
convergence---this allows it, for instance, to capture the optimal sparsity
pattern \citep[Ch.~16.4 \& 16.5]{nocedalwright}.
\citet{vinyes} provide a more in-depth discussion of the connections between
the two algorithms. We perform an empirical
comparison on a dependency parsing instance with random potentials.
Figure~\ref{fig:solver} shows that active set substantially outperforms all CG
variants, both in terms of objective value as well as in the solution sparsity,
suggesting that the quadratic curvature makes \smap solvable in very few
iterations to high accuracy.
We therefore use the active set solver in the remainder of the paper.

\subsection{Backpropagating Gradients through {\boldmath \smap}}
\label{sec:backpass}

In order to use {\smap} as a neural network layer trained with backpropagation,
one must compute products of the {\smap} Jacobian with a vector $\bs{p}$.
Computing the Jacobian of an optimization problem is an active research topic
known as {\em argmin differentiation}, and is generally difficult.
Fortunately, as we show next, argmin differentiation is always easy and
efficient in the case of {\smap}.

\begin{proposition}{\label{prop:backw}}
Denote a {\smap} solution by $\bs{y}^\star$
and its support by $\supp\coloneqq\{s~:~y_s > 0 \}$. Then, {\smap}
is differentiable almost everywhere with Jacobian
\[
    \begin{aligned}
        \pfrac{\bs{u}^\star}{\pr} &= \bs{M} \bs{D}(\supp) \bs{A}^\tr,
    ~\text{where}~\bs{D}(\supp) = \bs{D}(\supp)^\tr \text{given by}\\
    \bs{d}(\supp)_s &\coloneqq \begin{cases}
        \left(\bs{I} - \frac{1}{\bs{1}^T \bs{Z} \bs{1}}
        \bs{Z}\bs{1}\bs{1}^T\right) \col{Z}{s}, & s \in \supp \\
        \bs{0} & s \notin \supp \\
    \end{cases},
    \\
    \bs{Z} &\coloneqq
    ({\bs{M}_\supp}^\tr
    \bs{M}_\supp)^{-1}. \\
\end{aligned}
\]
\end{proposition}
The proof, given in Appendix~\ref{supp:jacobian}, relies on the KKT conditions
of the {\smap} QP. Importantly, because $\bs{D}(\mathcal{I})$ is zero outside
of the support of the solution, computing the Jacobian only requires the
columns of $\bs{M}$ and $\bs{A}$ corresponding to the structures in the active
set.  Moreover, when using the active set algorithm discussed in
\secref{solve}, the matrix $\bs{Z}$ is readily available as a byproduct of the
forward pass.  The backward pass can, therefore, be computed in
$\mathcal{O}(k|\mathcal{I}|)$.

Our approach for gradient computation draws its efficiency from the solution
sparsity and does not depend on the type of structure considered.  This is
contrasted with two related lines of research.  The first is  ``unrolling''
iterative inference algorithms, for instance belief propagation~\cite{ves,domke}
and gradient descent~\cite{spen}, where the backward pass complexity scales with
the number of iterations. In the second, employed by \citet{rush},
when inference can be performed via dynamic programming, backpropagation can be
performed using second-order expectation semirings \citep{Li2009} or more
general smoothing \cite{arthurdp},
in the same time complexity as the forward pass.
Moreover, in our approach, neither the forward nor the backward passes involve
logarithms, exponentiations or log-domain classes, avoiding the slowdown and
stability issues normally incurred. 

In the unstructured case, since $\bs{M}=\bs{I}$, $\bs{Z}$ is also an identity
matrix, uncovering the $\sparsemax$ Jacobian \citep{sparsemax}.
In general, structures are not necessarily orthogonal, but may have degrees of overlap.

\section{Structured Fenchel-Young Losses \\ and the {\boldmath \smap} Loss}
\label{sec:loss}

With the efficient algorithms derived above in hand, we switch gears to
defining a {\smap} {\em loss function}.
Structured output prediction models are typically trained by minimizing a {\em
structured loss} measuring the discrepancy between the desired structure
(encoded, for instance, as an indicator vector $\bs{y}=\bs{e}_s$) and the
prediction induced by the log-potentials $\pr$. 
We provide here a general family of structured prediction losses that will make
the newly proposed {\smap} loss arise as a very natural case.  Below, we let
$\Omega:\mathbb{R}^D \rightarrow \mathbb{R}$ denote a convex penalty function
and denote by $\Omega_\Simplex$ its restriction to $\Simplex^D \subset
\mathbb{R}^D$, \ie,
\[
    \Omega_\Simplex(\bs{y}) \coloneqq \begin{cases}
        \Omega(\bs{y}), & \bs{y} \in \Simplex^D; \\
        \infty, & \bs{y} \notin \Simplex^D. \\
    \end{cases}
\]
The Fenchel convex conjugate of $\Omega_\Simplex$ is
\[
    \Omega_\Simplex^\star(\pt) \coloneqq \sup_{\bs{y} \in \mathbb{R}^D} \pt^\tr\bs{y} -
    \Omega_\Simplex(\bs{y}) = \sup_{\bs{y} \in \Simplex^D} \pt^\tr\bs{y} -
    \Omega(\bs{y}).
\]
We next introduce a family of structured prediction losses,
named after the corresponding Fenchel-Young duality gap.
\begin{definition}[Fenchel-Young losses] Given a convex penalty function
$\Omega : \mathbb{R}^D \rightarrow \mathbb{R}$, and a $({k}\times{D})$-dimensional
matrix $\bs{A}=[\bs{M};\bs{N}]$ encoding the structure of the problem, we define
the following family of structured losses:
\begin{equation}\label{eq:fyloss}
    \ell_{\Omega,\bs{A}}(\pr, \bs{y})\coloneqq
    \Omega_\Simplex^\star(\bs{A}^\tr\pr) + \Omega_\Simplex(\bs{y}) -
    \pr^\tr\bs{A}\bs{y}.
\end{equation}
\end{definition}
This family, studied in more detail in~\cite{fylosses},
includes the commonly-used structured losses:
\begin{itemizesquish}
    \item Structured perceptron~\cite{perceptron}: $\Omega \equiv 0$;
    \item Structured SVM~\cite{ssvm,ssvmts}: $\Omega \equiv \rho(\cdot, \bar{\bs{y}})$ for a cost function
        $\rho$, where $\bar{\bs{y}}$ is the true output;
        \item CRF~\cite{crf}: $\Omega \equiv -H$;
    \item Margin CRF~\cite{margincrf}:\\ $\Omega \equiv -H + \rho(\cdot, \bar{\bs{y}})$.
\end{itemizesquish}
This leads to a natural way of defining {\smap} losses, by plugging the following into Equation~\ref{eq:fyloss}:
\begin{itemizesquish}
    \item {\smap} loss: $\Omega(\bs{y}) = \frac{1}{2}\norm{\bs{My}}^2_2$,
    \item Margin {\smap}: $\Omega(\bs{y}) = \frac{1}{2}\norm{\bs{My}}^2_2
        + \rho(\bs{y}, \bar{\bs{y}})$.
\end{itemizesquish}
It is well-known that the subgradients of structured perceptron and SVM losses
consist of MAP inference, while the CRF loss gradient requires marginal
inference.  Similarly, the subgradients of the {\smap} loss can be computed via
{\smap} inference, which in turn only requires MAP. The next proposition states
properties of structured Fenchel-Young losses, including a general connection
between a loss and its corresponding inference method.

\begin{proposition}
\label{prop:omega}
Consider a convex $\Omega$ and
a structured model defined by the matrix $\bs{A} \in \mathbb{R}^{k \times D}$.
Denote the inference objective $f_\Omega(\bs{y})\coloneqq\pr^\tr\bs{Ay} -
\Omega(\bs{y})$, and a solution $\bs{y}^\star \coloneqq \displaystyle
\argmax_{\bs{y} \in \Simplex^D} f_\Omega(\bs{y})$. Then, the following
properties hold:
\begin{enumeratesquish}
\item $\ell_{\Omega,\bs{A}}(\pr, \bs{y}) \geq 0$,
    with equality when $f_\Omega(\bs{y}) = f_\Omega(\bs{y}^\star)$;

\item $\ell_{\Omega,\bs{A}}(\pr, \bs{y})$ is convex, 
    $\partial \ell_{\Omega,\bs{A}}(\pr, \bs{y}) \ni \bs{A}(\bs{y}^\star -
    \bs{y})$;
\item $\ell_{t\Omega,\bs{A}}(\pr, \bs{y}) = t\ell_{\Omega}(\pr / t, \bs{y})$ for any $t \in \mathbb{R}, t>0$.
\end{enumeratesquish}
\end{proposition}
Proof is given in Appendix~\ref{supp:loss}.  Property 1 suggests that
pminimizing $\ell_{\Omega,\bs{A}}$ aligns models with the true label.
Property 2 shows how to compute subgradients of $\ell_{\Omega, \bs{A}}$
provided access to the inference output $[\bs{u}^\star; \bs{v}^\star] =
\bs{Ay^\star} \in \mathbb{R}^k$.  Combined with our efficient procedure
described in Section~\ref{sec:solve}, it makes the {\smap} losses promising for
structured prediction.  Property 3 suggests that the strength of the penalty
$\Omega$ can be adjusted by simply scaling $\pr$.  Finally, we remark that for
a strongly-convex $\Omega$, $\ell_{\Omega,\bs{A}}$ can be seen as a smoothed
perceptron loss; other smoothed losses have been explored by \citet{proxsdca}.

\section{Experimental Results}
\label{sec:exp}
In this section, we experimentally validate {\smap} on two natural language
processing applications, illustrating the two main use cases presented:
structured output prediction with the {\smap} loss~(\secref{parse})
and structured hidden layers~(\secref{esim}).
All models are implemented using the {\tt dynet} library v2.0.2~\cite{dynet}.
\subsection{Dependency Parsing with the {\boldmath \smap} Loss}
\label{sec:parse}
\begin{table}[t]
    \caption{\label{tab:parse}Unlabeled attachment accuracy scores for
        dependency parsing, using a bi-LSTM model~\cite{kg}.
        {\smap} and its margin version,
        m-{\smap}, produce the best parser on 4/5 datasets.
        For context, we include the scores of the
        CoNLL 2017 UDPipe baseline, which is trained under the same
        conditions~\cite{udpipe}.
}
    \small
    \begin{tabular}{r c c c c c}
        \toprule
        Loss & en & zh & vi & ro & ja \\
        \midrule
        Structured SVM &     87.02 &      81.94 &     69.42 &     87.58 &{\bf 96.24}\\
        CRF            &     86.74 &      83.18 &     69.10 &     87.13 &     96.09 \\
        {\smap}        &     86.90 & {\bf 84.03}&     69.71 &     87.35 &     96.04 \\
        m-{\smap}      &{\bf 87.34}&      82.63 &{\bf 70.87}&{\bf 87.63}&     96.03 \\
        \midrule
        UDPipe baseline& 87.68 & 82.14 & 69.63 & 87.36 & 95.94 \\
        \bottomrule
    \end{tabular}
\end{table}
\begin{figure*}[t]
    \small
    \centering
\newtoggle{squeeze}
\toggletrue{squeeze}
\begin{dependency}[hide label,edge unit distance=1.5ex,thick,label style={scale=1.3}]
\begin{deptext}
$\star$ \& They \& did \& a \& vehicle \& wrap \& for \& my \& Toyota \& Venza \& that \& looks \& amazing \& .\\
\end{deptext}
\depedge{3}{2}{1.0}
\depedge{1}{3}{1.0}
\depedge{6}{4}{1.0}
\depedge{6}{5}{1.0}
\depedge{3}{6}{1.0}
\depedge{10}{7}{1.0}
\depedge{10}{8}{1.0}
\depedge{10}{9}{1.0}
\depedge[very thick,color=myblue,show label,edge below,edge unit distance=0.5ex]{3}{10}{.55}
\depedge[very thick, color=myblue,show label,edge above,edge unit distance=1.5ex]{6}{10}{.45}
\depedge{12}{11}{1.0}
\depedge[very thick,color=mypurple,show label,edge below,edge unit distance=0.7ex]{3}{12}{.68}
\depedge[very thick,color=mypurple,show label,edge above,edge unit
distance=1.5ex]{6}{12}{.32}
\depedge{12}{13}{1.0}
\end{dependency}
\iftoggle{squeeze}{
\vspace{-0.5cm}
\raisebox{0.38cm}{
\begin{dependency}[hide label,edge unit distance=1.5ex,thick,label style={scale=1.3}]
\begin{deptext}
$\star$
\& the
\& broccoli
\& looks
\& browned
\& around
\& the
\& edges
\& . 
\\
\end{deptext}
\depedge{3}{2}{}
\depedge{4}{3}{}
\depedge[segmented edge,edge unit distance=1.3ex]{1}{4}{}
\depedge{4}{5}{}
\depedge{8}{6}{}
\depedge{8}{7}{}
\depedge[very thick,show label,edge below,color=myblue, edge unit distance=0.9ex]{4}{8}{.76}
\depedge[very thick,show label,edge above,color=myblue,edge unit distance=1.8ex]{5}{8}{.24}
\end{dependency}
}
}{
\\[0.5cm]
\begin{dependency}[hide label,edge unit distance=1.5ex,thick,label style={scale=1.3}]
\begin{deptext}
*
\& Salad
\& bar
\& is
\& hit
\& and
\& miss
\& for
\& freshness
\& --
\& sometimes
\& the
\& broccoli
\& looks
\& browned
\& around
\& the
\& edges
\& . 
\\
\end{deptext}
\depedge{3}{2}{}
\depedge{5}{3}{}
\depedge{5}{4}{}
\depedge{1}{5}{}
\depedge{7}{6}{}
\depedge{5}{7}{}
\depedge{9}{8}{}
\depedge[edge unit distance=1.2ex]{5}{9}{}
\depedge{5}{10}{}
\depedge{14}{11}{}
\depedge{13}{12}{}
\depedge{14}{13}{}
\depedge[segmented edge,edge unit distance=1.2ex]{5}{14}{}
\depedge{14}{15}{}
\depedge{18}{16}{}
\depedge{18}{17}{}
\depedge[very thick,show label,edge below,color=myblue, edge unit distance=0.9ex]{14}{18}{.76}
\depedge[very thick,show label,edge above,color=myblue,edge unit distance=1.8ex]{15}{18}{.24}
\end{dependency}
}
     \caption{\label{fig:ambiparse}Example of ambiguous parses from the
        UD English validation set. {\smap} selects a small number of candidate
        parses (left: three, right: two), differing from each other in a small
        number of ambiguous dependency arcs. In both cases, the desired gold
        parse is among the selected trees (depicted by the arcs above the
        sentence), but it is not the highest-scoring one.}
\end{figure*}
\begin{figure*}[t]
    \centering
    \includegraphics[width=0.99\textwidth]{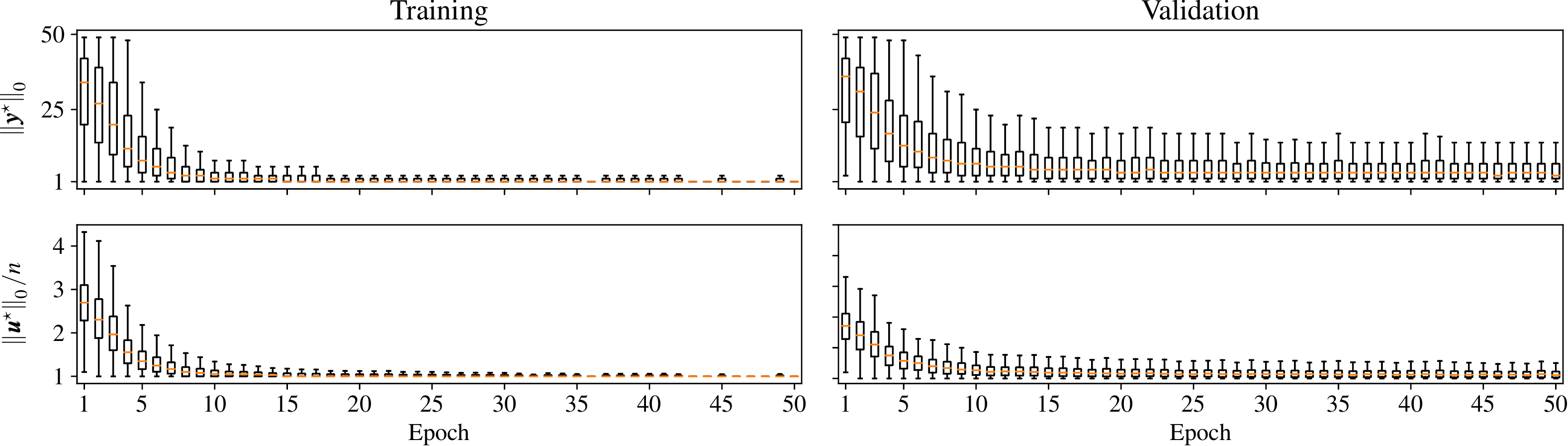}
    \caption{\label{fig:parsesparse}Distribution of the tree sparsity (top) and
    arc sparsity (bottom) of {\smap} solutions during training on the Chinese dataset. Shown are respectively the number of trees and the average number of parents per word with nonzero probability.}
\end{figure*}

We evaluate the {\smap} losses
against the commonly used CRF and structured SVM losses.
The task we focus on is non-projective {\em dependency parsing}: a structured output
task consisting of predicting the directed tree of
grammatical dependencies between words in a sentence~\citep[Ch.~14]{jurafsky-martin}.
We use annotated Universal Dependency data~\cite{ud},
as used in the CoNLL 2017 shared task~\cite{conll17}. 
To isolate the effect of the loss, we use the provided gold
tokenization and part-of-speech tags.
We follow closely the bidirectional LSTM arc-factored parser of
\citet{kg}, using the same model configuration;
the only exception is not using externally pretrained embeddings.
Parameters are trained using Adam \cite{adam}, tuning the
learning rate on the grid $\{.5, 1, 2, 4, 8\} \times 10^{-3}$,
expanded by a factor of 2 if the best model is at either end.

We experiment with 5 languages, diverse both in terms of family and in
terms of the amount of training data (ranging from 1,400 sentences
for Vietnamese to 12,525 for English). 
Test set results (Table~\ref{tab:parse}) indicate that the {\smap}
losses outperform the SVM and CRF losses on 4 out of the 5 languages
considered.  This suggests that {\smap} is a good middle ground between
MAP-based and marginal-based losses in terms of smoothness and gradient sparsity.

Moreover, as illustrated in Figure~\ref{fig:parsesparse}, the {\smap} loss
encourages {\bf sparse predictions}: models converge towards sparser solutions
as they train, yielding very few ambiguous arcs.  When confident, {\smap} can
predict a single tree.  Otherwise, the small set of candidate parses returned
can be easily visualized, often indicating genuine linguistic ambiguities
(Figure~\ref{fig:ambiparse}). Returning a small set of parses, also sought 
concomittantly by \citet{ocon}, is valuable in pipeline systems, \eg, when the
parse is an input to a downstream application: error propagation is diminished in cases
where the highest-scoring tree is incorrect (which is the case for the
sentences in Figure~\ref{fig:ambiparse}). Unlike $K$-best heuristics, {\smap}
dynamically adjusts its output sparsity, which is desirable on realistic data
where most instances are easy.

\subsection{Latent Structured Alignment\\for Natural Language Inference}
\label{sec:esim}
\begin{figure*}
    \subcaptionbox{softmax}{%
    \includegraphics[width=0.33\textwidth]{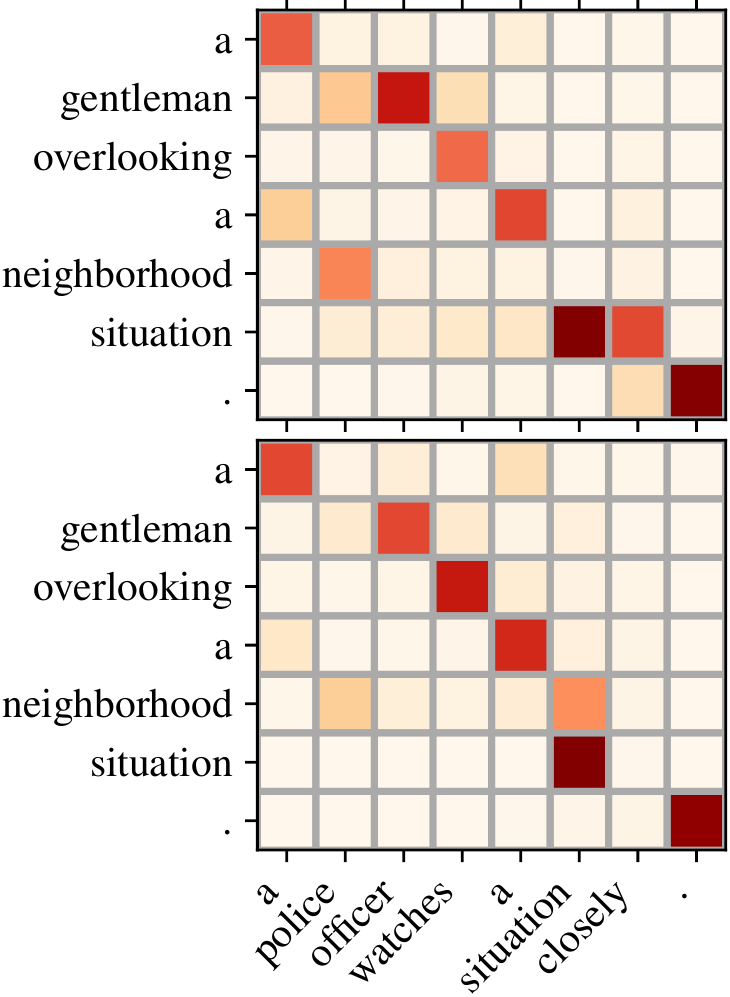}}
    \subcaptionbox{sequence}{%
    \includegraphics[width=0.33\textwidth]{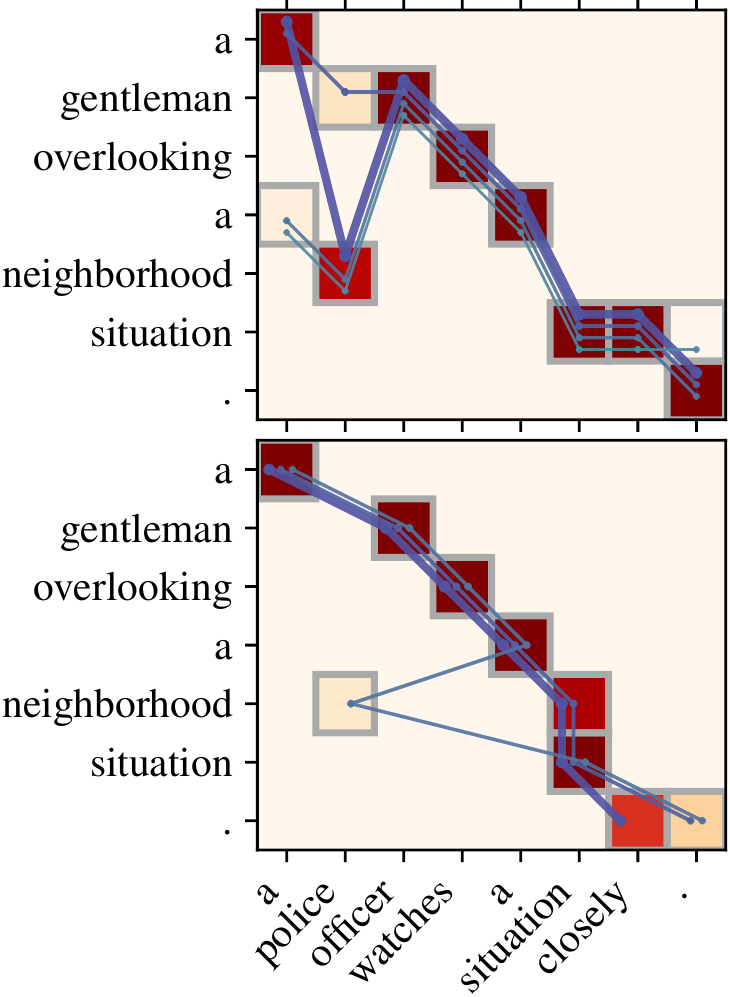}}
    \subcaptionbox{matching}{%
      \vbox{%
        \hbox{%
\begin{tikzpicture}
\small
\def\stretch{2.3pt};
\node[align=right] at (0,0){%
a\\[\stretch]
gentleman\\[\stretch]
overlooking\\[\stretch]
a\\[\stretch]
neighborhood\\[\stretch]
situation\\[\stretch]
.%
};

\node[align=left] at (4,0){%
a\\[\stretch]
police\\[\stretch]
officer\\[\stretch]
watches\\[\stretch]
a\\[\stretch]
situation\\[\stretch]
closely\\[\stretch]
.%
};

\def\xp{1.05};
\def\base{-1.17};
\def\H{0.425};
\coordinate (p1) at (\xp, \base+(6*\H););
\coordinate (p2) at (\xp, \base+(5*\H););
\coordinate (p3) at (\xp, \base+(4*\H););
\coordinate (p4) at (\xp, \base+(3*\H););
\coordinate (p5) at (\xp, \base+(2*\H););
\coordinate (p6) at (\xp, \base+\H);
\coordinate (p7) at (\xp, \base);

\def\xh{3.25};
\def\baseh{-1.38};
\coordinate (h1) at (\xh, \baseh+(7*\H););
\coordinate (h2) at (\xh, \baseh+(6*\H););
\coordinate (h3) at (\xh, \baseh+(5*\H););
\coordinate (h4) at (\xh, \baseh+(4*\H););
\coordinate (h5) at (\xh, \baseh+(3*\H););
\coordinate (h6) at (\xh, \baseh+(2*\H););
\coordinate (h7) at (\xh, \baseh+\H);
\coordinate (h8) at (\xh, \baseh);

\def\OFS{1.6pt};
\def\LB{0.3pt};
\def\LW{1.7pt};

\begin{scope}[transform canvas={yshift=-4*\OFS},tab20n4,line width=\LB+0.02*\LW]
    \draw (p1) -- (h5);
    \draw (p2) -- (h2);
    \draw (p3) -- (h4);
    \draw (p4) -- (h1);
    \draw (p5) -- (h3);
    \draw (p6) -- (h6);
    \draw (p7) -- (h8);
\end{scope}
\begin{scope}[transform canvas={yshift=-3*\OFS},tab20n3,line width=\LB+0.05*\LW]
    \draw (p1) -- (h3);
    \draw (p2) -- (h1);
    \draw (p3) -- (h4);
    \draw (p4) -- (h5);
    \draw (p5) -- (h2);
    \draw (p6) -- (h6);
    \draw (p7) -- (h8);
\end{scope}
\begin{scope}[transform canvas={yshift=-2*\OFS},tab20n2,line width=\LB+0.05*\LW]
    \draw (p1) -- (h1);
    \draw (p2) -- (h3);
    \draw (p3) -- (h4);
    \draw (p4) -- (h5);
    \draw (p5) -- (h2);
    \draw (p6) -- (h6);
    \draw (p7) -- (h8);
\end{scope}
\begin{scope}[transform canvas={yshift=-1*\OFS},tab20n1,line width=\LB+0.30*\LW]
    \draw (p1) -- (h1);
    \draw (p2) -- (h2);
    \draw (p3) -- (h4);
    \draw (p4) -- (h5);
    \draw (p5) -- (h3);
    \draw (p6) -- (h6);
    \draw (p7) -- (h8);
\end{scope}
\begin{scope}[tab20n0,line width=\LB + 0.57*\LW]
    \draw (p1) -- (h1);
    \draw (p2) -- (h3);
    \draw (p3) -- (h4);
    \draw (p4) -- (h5);
    \draw (p5) -- (h2);
    \draw (p6) -- (h6);
    \draw (p7) -- (h7);
\end{scope}
\end{tikzpicture}
}%
        \hbox{\includegraphics[width=0.33\textwidth]{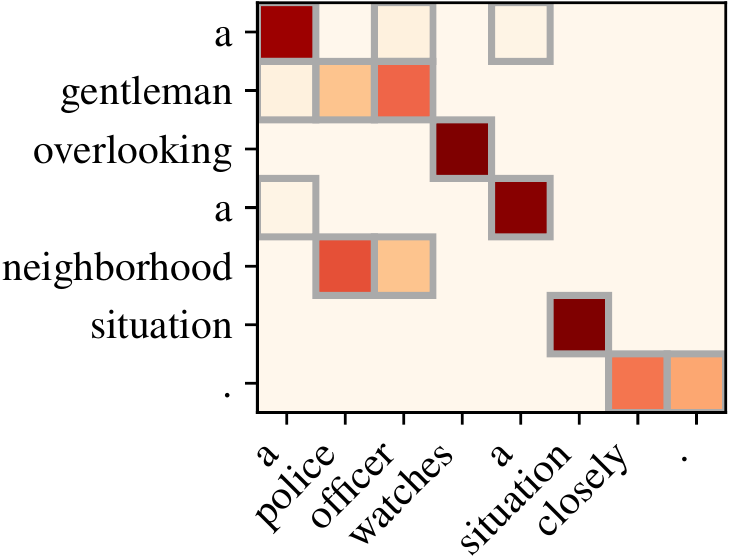}}%
      }
    }
    \caption{\label{fig:snli}Latent alignments on an example from the SNLI
    validation set, correctly predicted as {\em neutral} by all compared
    models. The premise is on the $y$-axis, the hypothesis on the
    $x$-axis. Top: columns sum to 1;
    bottom: rows sum to 1. The matching alignment mechanism yields a
    symmetrical alignment, and is thus shown only once. Softmax yields
    a dense alignment (nonzero weights are marked with a border).
    The structures selected by sequential alignment are overlayed
    as paths; the selected matchings are displayed in the top right.}
\end{figure*}

\begin{table}[t]
\caption{\label{tab:nli}Test accuracy scores for natural language inference with
structured and unstructured variants of ESIM. In parentheses: the percentage of pairs of words
with nonzero alignment scores.}
    \centering
    \small
    \begin{tabular}{r c@{ }r c@{ }r}
    \toprule
    ESIM variant & \multicolumn{2}{c}{MultiNLI} & \multicolumn{2}{c}{SNLI} \\
    \midrule
    softmax      &     76.05 &{\footnotesize \color{supgray} (100\%)} &     86.52 &{\footnotesize \color{supgray}(100\%)}\\
    sequential   &     75.54 &{\footnotesize \color{supgray}(13\%)} &{\bf 86.62}  &{\footnotesize \color{supgray}(19\%)}\\
    matching     &{\bf 76.13}&{\footnotesize \color{supgray} (8\%)}&     86.05    &{\footnotesize \color{supgray}(15\%)}\\
    \bottomrule
    \end{tabular}
\end{table}

In this section, we demonstrate {\smap} for inferring latent
structure in large-scale deep neural networks. We focus on the task of {\em
natural language inference}, defined as the classification problem of deciding,
given two sentences (a {\em premise} and a {\em hypothesis}), whether the
premise {\em entails} the hypothesis, {\em contradicts} it, or is {\em neutral}
with respect to it.

We consider novel structured variants of the state-of-the-art ESIM model~\cite{esim}.
Given a premise $\mathsf{P}$ of length $m$ and a hypothesis $\mathsf{H}$ of length $n$,
ESIM:
\begin{enumeratesquish}
    \item Encodes $\mathsf{P}$ and $\mathsf{H}$ with an LSTM.
    \item Computes alignment scores $\bs{G}\in\mathbb{R}^{m\times n}$;
    with $g_{ij}$ the inner product between the $\mathsf{P}$ word $i$
    and $\mathsf{H}$ word $j$. 
    \item Computes $\mathsf{P}$-to-$\mathsf{H}$ and $\mathsf{H}$-to-$\mathsf{P}$ alignments
    using row-wise, respectively column-wise $\softmax$ on $\bs{G}$.
    \item Augments $\mathsf{P}$ words with the weighted average of its aligned $\mathsf{H}$
    words, and vice-versa.
    \item Passes the result through another LSTM, then predicts.
\end{enumeratesquish}

We consider the following structured replacements for the independent row-wise
and column-wise $\softmax$es (step 3):

{\bf Sequential alignment.}
We model the alignment of $\bs{p}$ to
$\bs{h}$ as a sequence tagging instance of length $m$, with $n$ possible
tags corresponding to the $n$ words of the hypothesis. Through {\em
transition scores}, we enable the model to capture continuity and monotonicity
of alignments: we parametrize transitioning from word $t_1$ to $t_2$
by binning the distance $t_2 - t_1$ into 5 groups,
$\{-2\text{ or less}, -1, 0, 1, 2 \text{ or more}\}$. We similarly parametrize the initial alignment
using bins $\{1, 2 \text{ or more}\}$ and the final alignment as $\{-2
\text{ or less}, -1\}$, allowing the model to express whether an alignment starts at the
beginning or ends on the final word of $\bs{h}$; formally
\[
    \eta_F(i, t_1, t_2) \coloneqq \begin{cases}
        w_{\text{bin}(t_2 - t_1)} & 0 < i < n, \\
        w^{\text{start}}_{\text{bin}(t_2)} & i = 0, \\
        w^{\text{end}}_{\text{bin}(t_1)} & i = n. \\
    \end{cases}
\]
We align $\bs{p}$ to $\bs{h}$ applying the same method in the other
direction, with different transition scores $\bs{w}$. Overall, sequential
alignment requires learning 18 additional scalar parameters.

{\bf Matching alignment.}
We now seek a symmetrical alignment in both directions simultaneously. To this
end, we cast the alignment problem as finding a maximal weight bipartite
matching.  We recall from \secref{structinf} that a solution can be found via
the Hungarian algorithm (in contrast to marginal inference, which is
\#P-complete). When $n=m$, maximal matchings can be represented as permutation
matrices, and when $n\neq m$ some words remain unaligned. {\smap} returns a
weighted average of a few maximal matchings. This method requires no additional
learned parameters.

We evaluate the two models alongside the $\softmax$ baseline on the
SNLI~\cite{snli} and MultiNLI~\cite{multinli} datasets.\footnote{We split the
MultiNLI matched validation set into equal validation and test sets; for SNLI
we use the provided split.} All models are trained by SGD, with $0.9\times$
learning rate decay at epochs when the validation accuracy is not the best
seen. We tune the learning rate on the grid $\big\{2^k : k \in \{-6, -5, -4,
-3\} \big\}$, extending the range if the best model is at either end.
The results in Table~\ref{tab:nli} show that structured alignments are
competitive with $\softmax$ in terms of accuracy, but are orders of magnitude
sparser. This sparsity allows them to produce global alignment structures that
are interpretable, as illustrated in Figure~\ref{fig:snli}.

Interestingly, we observe computational advantages of sparsity. 
Despite the overhead of GPU memory copying, both training and validation in our
latent structure models take roughly the same time as with
$\softmax$ and become faster as the models grow more certain. For
the sake of comparison, \citet{rush} report a $5\times$ slow-down in their structured
attention networks, where they use marginal inference.

\section{Related Work}
{\bf Structured attention networks.}
\citet{rush} and \citet{lapata} take advantage of the tractability of marginal
inference in certain structured models and derive specialized backward passes
for structured attention.
In contrast, our approach is modular and general: with {\smap}, the forward pass
only requires MAP inference, and the backward pass is efficiently computed
based on the forward pass results.  Moreover, unlike marginal inference,
{\smap} yields sparse solutions, which is an appealing property statistically,
computationally, and visually.

{\boldmath \bf $K$-best inference.}
As it returns a small set of structures, {\smap} brings to mind $K$-best
inference, often used in pipeline NLP systems for increasing recall and
handling uncertainty~\cite{bishan}. $K$-best inference can be approximated
(or, in some cases, solved), roughly $K$ times slower than MAP inference
\cite{kbest,kbesttree,kbestmatch,kbestlp}. 
The main advantages of \smap are convexity, differentiablity, and modularity,
as \smap can be computed \emph{in terms of} MAP subproblems.
Moreover, it yields a distribution,
unlike $K$-best, which does not reveal the gap between selected structures,

\textbf{Learning permutations.} A popular approach for differentiable
permutation learning involves mean-entropic optimal transport relaxations
\cite{sinkprop,mena}. Unlike \smap, this does not apply to general structures,
and solutions are not directly expressible as combinations of a few permutations.

{\bf Regularized inference.}
\citet{Ravikumar2010}, \citet{meshi2015smooth}, and~\citet{ad3} proposed
$\ell_2$ perturbations and penalties in various related ways, with the goal of
solving LP-MAP approximate inference in graphical models.  In contrast, the
goal of our work is sparse structured prediction, which is not considered in
the aforementioned work. Nevertheless, some of the formulations in their work
share properties with  {\smap}; exploring the connections further is an
interesting avenue for future work.

\section{Conclusion}
We introduced a new framework for sparse structured inference, {\smap}, along
with a corresponding loss function. We proposed efficient ways to compute the
forward and backward passes of {\smap}. Experimental results illustrate two use
cases where sparse inference is well-suited. For structured prediction, the
{\smap} loss leads to strong models that make sparse, interpretable
predictions, a good fit for tasks where local ambiguities are common, like many
natural language processing tasks. For structured hidden layers, we
demonstrated that {\smap} leads to strong, interpretable networks trained
end-to-end. Modular by design, {\smap} can be applied readily to any structured
problem for which MAP inference is available, including combinatorial problems
such as linear assignment.
\section*{Acknowledgements}
We thank
Tim Vieira,
David Belanger,
Jack Hessel,
Justine Zhang,
Sydney Zink,
the Unbabel AI Research team,
and the three anonymous reviewers for their insightful 
comments.
This work was %
supported by the European Research Council (ERC
StG DeepSPIN 758969) 
and by the Funda\c{c}\~ao para a Ci\^encia e Tecnologia through contracts UID/EEA/50008/2013, PTDC/EEI-SII/7092/2014 (LearnBig), and CMUPERI/TIC/0046/2014 ~(GoLocal).
\vspace{-10pt}
\setlength{\bibsep}{.8ex plus 0.2ex}

\bibliographystyle{icml2018}

\clearpage
\onecolumn
\appendix
\begin{center}
{\Huge \bf Supplementary material}
\end{center}
\section{Implementation Details for {\boldmath \smap} Solvers}
\label{supp:solvers}
\subsection{Conditional Gradient Variants}
We adapt the presentation of vanilla, away-step and pairwise conditional gradient
of \citet{cg}.

Recall the {\smap} optimization problem (Equation~\ref{eqn:smap}), which we
rewrite below as a minimization, to align with the formulation in~\cite{cg}
\[
    \smapop_{\bs{A}}(\pr)\coloneqq
    \argmin_{\bs{u}:~[\bs{u}, \bs{v}] \in \mathcal{M}_{\bs{A}}}
    f(\bs{u}, \bs{v}), \qquad \text{where}~f(\bs{u},\bs{v})\coloneqq
    \frac{1}{2} \norm{\bs{u}}_2^2
    -\pr_U^\tr\bs{u} - \pr_F^\tr \bs{v}.
\]
The gradients of the objective function $f$  w.r.t.\ the two variables are
\[
    \nabla_{\bs{u}} f(\bs{u}', \bs{v}') = \bs{u}' - \pr_U, \qquad 
    \nabla_{\bs{v}} f(\bs{u}', \bs{v}') = - \pr_V.
\]
The ingredients required to apply conditional gradient algorithms are solving
linear minimization problem, selecting the away step, computing the Wolfe gap,
and performing line search.

\paragraph{Linear minimization problem.} For {\smap}, this amounts to a MAP
inference call, since
\[
\begin{aligned}
     & \argmin_{[\bs{u}, \bs{v}] \in \mathcal{M}_{\bs{A}}}
    ~ \bigl\langle \nabla_{\bs{u}} f(\bs{u'}, \bs{v'}), \bs{u} \bigr\rangle
    + \bigl\langle \nabla_{\bs{v}} f(\bs{u'}, \bs{v'}), \bs{v} \bigr\rangle\\
    =& \argmin_{[\bs{u}, \bs{v}] \in \mathcal{M}_{\bs{A}}}
    ~ (\bs{u}' - \pr_U)^\tr \bs{u} - \pr_F^\tr\bs{v} \\ 
    =&~\{[\col{M}{s}; \col{N}{s}]~:~ s \in \map_{\bs{A}}(\pr_U - \bs{u}',
    \pr_F)\}.
\end{aligned}
\]
where we assume $\map_{\bs{A}}$ yields the set of maximally-scoring structures.

\paragraph{Away step selection.} This step involves searching the currently
selected structures in the active set $\mathcal{I}$ with the {\em opposite}
goal: finding the structure {\em maximizing} the linearization
\[
\begin{aligned}
    & \argmax_{s \in \mathcal{I}}
    ~ \bigl\langle \nabla_{\bs{u}} f(\bs{u'}, \bs{v'}), \col{M}{s} \bigr\rangle
    + \bigl\langle \nabla_{\bs{v}} f(\bs{u'}, \bs{v'}), \col{N}{s} \bigr\rangle\\
    =& \argmax_{s \in \mathcal{I}}
    ~ (\bs{u}' - \pr_U)^\tr \col{M}{s} - \pr_F^\tr\col{N}{s} \\ 
\end{aligned}
\]

\paragraph{Wolfe gap.}
The gap at a point $\bs{d} = [\bs{d}_{\bs{u}}; \bs{d}_{\bs{v}}]$ is given by
\begin{equation}
\label{eqn:gap}
\begin{aligned}
    \operatorname{gap}(\bs{d}, \bs{u}') &\coloneqq
        \bigl\langle -\nabla_{\bs{u}} f(\bs{u'}, \bs{v'}), \bs{d}_{\bs{u}} \bigr\rangle
        + \bigl\langle -\nabla_{\bs{v}} f(\bs{u'}, \bs{v'}), \bs{d}_{\bs{v}} \bigr\rangle\\
        &= \bigl \langle \pr_U - \bs{u}', \bs{d}_{\bs{u}}\bigr\rangle +
           \bigl \langle \pr_F, \bs{d}_{\bs{v}}\bigr\rangle.\\
\end{aligned}
\end{equation}

\paragraph{Line search.}
Once we have picked a direction $\bs{d}=[\bs{d}_{\bs{u}}; \bs{d}_{\bs{v}}]$, we can pick
the optimal step size by solving a simple optimization problem.
Let $\bs{u}_\gamma \coloneqq  \bs{u}' + \gamma\bs{d}_{\bs{u}}$,
and $\bs{v}_\gamma \coloneqq  \bs{v}' + \gamma\bs{d}_{\bs{v}}$.
We seek $\gamma$ so as to optimize
\[
    \argmin_{\gamma \in [0, \gamma_{\max}]} f(\bs{u}_\gamma,
    \bs{v}_\gamma)
\]
Setting the gradient w.r.t.\ $\gamma$ to $0$ yields
\[
\begin{aligned}
0 &= \pfrac{}{\gamma} f(\bs{u}_\gamma, \bs{v}_\gamma)\\
  &= \bigl \langle \bs{d}_{\bs{u}},\nabla_{\bs{u}} f(\bs{u}_\gamma, \bs{v}_\gamma)
     \bigr \rangle +
     \bigl \langle \bs{d}_{\bs{v}},\nabla_{\bs{v}} f(\bs{u}_\gamma,\bs{v}_\gamma)
     \bigr \rangle \\
  &= \bigl \langle \bs{d}_{\bs{u}}, \bs{u}' + \gamma \bs{d}_{\bs{u}} - \pr_U
     \bigr \rangle + \bigl \langle \bs{d}_{\bs{v}}, -\pr_F \bigr \rangle \\
  &= \gamma \norm{\bs{d}_{\bs{u}}}_2^2 + \bs{u}'^\tr\bs{d}_{\bs{u}}-\pr^\tr\bs{d}
\end{aligned}
\]
We may therefore compute the optimal step size $\gamma$ as
\begin{equation}
\label{eqn:linesearch}
    \gamma = \max\left(0, \min\left(\gamma_{\max}, 
        \frac{\pr^\tr \bs{d} - \bs{u}'^\tr \bs{d}_{\bs{u}}}{\norm{\bs{d}_{\bs{u}}}_2^2}
    \right ) \right )
\end{equation}

\begin{algorithm*}[h]
\caption{Conditional gradient for {\smap}}
\begin{algorithmic}[1]
    \STATE 
        Initialization:\quad
        $s^{(0)} \leftarrow \map_{\bs{A}}(\pr_U, \pr_F);
        \quad \mathcal{I}^{(0)}=\{s^{(0)}\};
        \quad \bs{y}^{(0)}=\bs{e}_{s^{(0)}};
        \quad [\bs{u}^{(0)};\bs{v}^{(0)}]=\col{A}{s^{(0)}}$

    \FOR{$t=0\dots t_{\max}$}

        \STATE $s\leftarrow\map_{\bs{A}}(\pr_U-\bs{u}^{(t)},\pr_F)$;
        \hskip7.7em
        $\bs{d}^{\text{F}} \leftarrow \col{A}{s}-[\bs{u}^{(t)};\bs{v}^{(t)}]$
        \quad(forward direction)

        \STATE $w \leftarrow \displaystyle\argmax_{w\in\mathcal{I}^{(t)}}~
        (\pr_U - \bs{u}^{(t)})^\tr \col{M}{w} + \pr_F^\tr \col{N}{w};
        \qquad \bs{d}^\text{W} \leftarrow
            [\bs{u}^{(t)}; \bs{v}^{(t)}] - \col{A}{w}$
        \quad(away direction)

        \IF {$\operatorname{gap}(\bs{d}^\text{F}, \bs{u}^{(t)}) < \epsilon$}
        \STATE{{\bf return} $\bs{u}^{(t)}$} \quad(Equation~\ref{eqn:gap})
        \ENDIF

        \IF {variant $=$ vanilla}
        \STATE $\bs{d} \leftarrow \bs{d}^\text{F}; \qquad \gamma_{\max}\leftarrow 1$
        \ELSIF{variant $=$ pairwise}
        \STATE $\bs{d} \leftarrow \bs{d}^\text{F}+\bs{d}^\text{W}; \qquad
        \gamma_{\max}\leftarrow y_w$
        \ELSIF{variant $=$ away-step}

        \IF{$
            \operatorname{gap}(\bs{d}^\text{F}, \bs{u}^{(t)}) \geq
            \operatorname{gap}(\bs{d}^\text{W}, \bs{u}^{(t)})$}
        
            \STATE $\bs{d} \leftarrow \bs{d}^\text{F}; \qquad \gamma_{\max}\leftarrow 1$
        \ELSE
            \STATE $\bs{d} \leftarrow \bs{d}^\text{A}; \qquad
            \gamma_{\max}\leftarrow {y_w}/{(1 - y_w)}$
        \ENDIF

        \ENDIF

        \STATE Compute step size $\gamma$
        \quad(Equation~\ref{eqn:linesearch})
        \STATE $[\bs{u}^{(t+1)}; \bs{v}^{(t+1)}] \leftarrow [\bs{u}^{(t)};
            \bs{v}^{(t)}] + \bs{d}$ 
        \STATE Update $\mathcal{I}^{(t+1)}$ and $\bs{y}^{(t+1)}$ accordingly.
    \ENDFOR
\end{algorithmic}
\end{algorithm*}

\subsection{The Active Set Algorithm}

We use a variant of the active set algorithm \citep[Ch.~16.4 \&
16.5]{nocedalwright} as proposed for the quadratic subproblems of the AD$^3$
algorithm; our presentation follows \citep[Algorithm 3]{ad3}.
At each step, the active set algorithm solves a relaxed variant of the {\smap}
QP, relaxing the non-negativity constraint on $\bs{y}$, and restricting the
solution to the current active set $\mathcal{I}$
\[
    \operatorname{minimize}_{\bs{y}_\supp \in \mathbb{R}^{|\supp|}}
    \quad\frac{1}{2}\norm{\bs{M}_{\supp}\bs{y}_{\supp}}_2^2 
        - \pr^\tr\bs{A}_{\supp}\bs{y}_{\supp} 
    \qquad\qquad \text{subject to} \quad \bs{1}^\tr\bs{y}_\supp = 1
\] 
whose solution can be found by solving the KKT system
\begin{equation}
    \label{eqn:relaxedqp}
    \begin{bmatrix}
        \bs{M}^\tr_\supp \bs{M}_\supp & \bs{1} \\
        \bs{1}^\tr & 0 \\
    \end{bmatrix}
    \begin{bmatrix} \bs{y}_\supp \\ \tau \end{bmatrix}
    = 
    \begin{bmatrix} \bs{A}^\tr_\supp \pr \\ 1 \end{bmatrix}.
\end{equation}
At each iteration, the (symmetric) design matrix in Equation~\ref{eqn:relaxedqp} is
updated by adding or removing a row and a column; therefore its inverse
(or a decomposition) may be efficiently maintained and updated.

\paragraph{Line search.}
The optimal step size for moving a feasible current estimate $\bs{y}'$
toward a solution $\hat{\bs{y}}$ of Equation~\ref{eqn:relaxedqp}, 
while keeping feasibility, is given by \citep[Equation 31]{ad3}
\begin{equation}
    \label{eqn:asgamma}
    \gamma = \min \left(1, \min_{s\in\supp,~y'_s>\hat{y}_s}
        \frac{y'_s}{y'_s - \hat{y}_s}
    \right)
\end{equation}
When $\gamma \leq 1$ this update zeros out a coordinate of $\bs{y}'$;
otherwise, $\supp$ remains the same.

\begin{algorithm*}[h]
\caption{Active Set algorithm for {\smap}}
\begin{algorithmic}[1]
    \STATE 
        Initialization:\quad $s^{(0)} \leftarrow \map_{\bs{A}}(\pr_U, \pr_F);
        \quad \mathcal{I}^{(0)}=\{s^{(0)}\};
        \quad \bs{y}^{(0)}=\bs{e}_{s^{(0)}};
        \quad [\bs{u}^{(0)};\bs{v}^{(0)}]=\col{A}{s^{(0)}}$

    \FOR{$t=0\dots t_{\max}$}

    \STATE Solve the relaxed QP restricted to $\mathcal{I}^{(t)}$;
    get $\hat{\bs{y}}, \hat{\tau}, \hat{\bs{u}}=\bs{M}\hat{\bs{y}}$
    \quad(Equation~\ref{eqn:relaxedqp})
    \IF {$\hat{\bs{y}} = \bs{y}^{(t)}$}
    \STATE $s \leftarrow \map_{\bs{A}}(\pr_U - \hat{\bs{u}},
    \pr_F)$

        \IF {$\operatorname{gap}(\col{A}{s}, \hat{\bs{u}})\leq \hat{\tau}$} 
        \STATE {\bf return} $\bs{u}^{(t)}$\quad(Equation~\ref{eqn:gap})
        \ELSE
        \STATE $\mathcal{I}^{(t+1)} \leftarrow \mathcal{I}^{(t)} \cup \{s\}$
        \ENDIF
    \ELSE
    \STATE Compute step size $\gamma$\quad(Equation~\ref{eqn:asgamma})
    \STATE $\bs{y}^{(t+1)} \leftarrow (1-\gamma)\bs{y}^{(t)} + \gamma
    \hat{\bs{y}}$\quad(sparse update)
    \STATE Update $\mathcal{S}^{(t+1)}$ if necessary
    \ENDIF

    \ENDFOR
\end{algorithmic}
\end{algorithm*}

\section{Computing the {\boldmath \smap} Jacobian: Proof of Proposition~\ref{prop:backw}}
\label{supp:jacobian}

Recall that {\smap} is defined as the $\bs{u}^\star$ that maximizes the value
of the quadratic program (Equation~\ref{eqn:smap}),

\begin{equation}
\label{eqn:qp_small}
g(\pr_U, \pr_F) \coloneqq \max_{[\bs{u};\bs{v}]\in\mathcal{M}_{\bs{A}}} \pr_U^\tr \bs{u} + \pr_F^\tr \bs{v}
-\frac{1}{2}\norm{\bs{u}}^2_2.
\end{equation}

As the $\ell_2^2$ norm is strongly convex, there is always a unique minimizer
$\bs{u}^\star$ (implying that $\smap$ is well-defined), and the convex conjugate
of the QP in (\ref{eqn:qp_small}), $g^*(\bs{u}, \bs{v})=\bigl\{ \frac{1}{2}\norm{\bs{u}}_2^2,
{[\bs{u};\bs{v}]}\in\mathcal{M}_{\bs{A}}; -\infty \text{ otherwise}\bigr \}$  is smooth in $\bs{u}$,
implying that $\smap$ (which only returns $\bs{u}$) is Lipschitz-continuous and thus
differentiable almost everywhere.

We now rewrite the QP in Equation~\ref{eqn:qp_small} in terms of the convex
combination of vertices of the marginal polytope

\begin{equation}
    \label{eq:regmin}
    \min_{\bs{y}\in\Simplex^D} \frac{1}{2}\norm{\bs{My}}^2_2 - \pt^\tr\bs{y} 
    \qquad \text{where}~
    \pt \coloneqq \bs{A}^\tr\pr
\end{equation}

We use the optimality conditions of problem \ref{eq:regmin} to derive an
explicit relationship between $\bs{u}^\star$ and $\bs{x}$.
At an optimum, the following KKT conditions hold

\begin{align}
    \bs{M}^\tr\bs{M}\bs{y}^\star - \bs{\lambda}^\star + \tau^\star\bs{1} &= \pt\\
    \bs{1}^\tr\bs{y}^\star &= 1 \\
    \bs{y}^\star &\geq \bs{0} \\
    \bs{\lambda}^\star &\geq \bs{0} \\ 
    \bs{\lambda}^{\star \tr}\bs{y}^\star &= 0 \label{eq:slack}
\end{align}

Let $\supp$ denote the support of $\bs{y}^\star$, \ie, $\supp = \{s~:~y^\star_s > 0 \}$.
From Equation~\ref{eq:slack} we have $\bs{\lambda}_\supp = \bs{0}$ and therefore
\begin{align}
    {\bs{M}_\supp}^\tr\bs{M}_\supp\bs{y}_\supp^\star + \tau^\star\bs{1} &=
    \pt_\supp \label{eq:sub1}\\
    \bs{1}^\tr\bs{y}_\supp^\star &= 1 \label{eq:sub2}
\end{align}
Solving for $\bs{y}_\supp^\star$ in Equation~\ref{eq:sub1} we get a direct expression 
\[
    {\bs{y}_\supp}^\star =
    ({\bs{M}_\supp}^\tr
    \bs{M}_\supp)^{-1}
    (\pt_\supp - \tau^\star \bs{1}) 
    =
    \bs{Z}(\pt_\supp - \tau^\star \bs{1}).
\]
where we introduced $\bs{Z}=(\bs{M}^\tr\bs{M})^{-1}$. Solving for $\tau^\star$
yields
\[
    \tau^\star = \frac{1}{\bs{1}^T \bs{Z} \bs{1}} \left(
    \bs{1}^T \bs{Z} \pt_\supp- 1 \right)\\
\]
Plugging this back and left-multiplying by $\bs{M}_\supp$ we get
\[
    \bs{u}^\star = 
    \bs{M}_\supp \bs{y}^\star_\supp = \bs{M}_\supp \bs{Z}\left(\pt_\supp - \frac{1}{\bs{1}^\tr \bs{Z1}} \bs{1}^\tr
    \bs{Z}\pt_\supp \bs{1} + \frac{1}{\bs{1}^\tr \bs{Z1}} \bs{1}\right)
\]
Note that, in a neighborhood of $\pr$, the support of the solution $\supp$ is
constant. (On the measure-zero set of points where the support changes, $\smap$
is subdifferentiable and our assumption yields a generalized
Jacobian~\cite{clarke_book}.)  Differentiating w.r.t.\ the score of a configuration $\theta_s$,
we get the expression
\begin{equation}
    \pfrac{\bs{u}^\star}{\theta_s} = 
    \begin{cases}
        \bs{M}\left(\bs{I} - \frac{1}{\bs{1}^T \bs{Z} \bs{1}}
        \bs{Z}\bs{1}\bs{1}^T\right) \col{Z}{s} & s \in \supp \\
        \bs{0} & s \notin \supp \\
    \end{cases}
\end{equation} 
Since $\theta_s = \col{A}{s}^\tr \pr$, by the chain rule, we get the desired
result
\begin{equation}
\pfrac{\bs{u}^\star}{\pr} = \pfrac{\bs{u}^\star}{\pt} \bs{A}^\tr.
\end{equation}

\section{Fenchel-Young Losses: Proof of Proposition~\ref{prop:omega}}
\label{supp:loss}

We recall that the structured Fenchel-Young loss defined by a convex
$\Omega:\mathbb{R}^D\rightarrow \mathbb{R}$ and a matrix $\bs{A}$ is defined as
\[
    \ell_{\Omega,\bs{A}}:\mathbb{R}^k \times \Simplex^D \rightarrow
    \mathbb{R},\quad
    \ell_{\Omega,\bs{A}}(\pr, \bs{y}) \coloneqq
    \Omega_\Simplex^*(\bs{A}^\tr\pr) + \Omega_\Simplex(\bs{y})-
    \pr^\tr\bs{A}\bs{y}.
\]
Since $\Omega_\Simplex$ is the restriction of a convex function to a convex set,
it is convex~\citep[Section 3.1.2]{boyd}.

\paragraph{Property 1.}
From the Fenchel-Young inequality~\citep[Section 3.3.2]{fenchel,boyd}, we have
\[
    \pt^\tr\bs{y} \leq \Omega_\Simplex^*(\pt) + \Omega_\Simplex(\bs{y}).
\]
In particular, when $\pt = \bs{A}^\tr \pr$, %
\[
    \begin{aligned}
        0 & \leq - \pr^\tr\bs{A}\bs{y} + 
         \Omega_\Simplex^*(\bs{A}^\tr \pr)
        + \Omega_\Simplex(\bs{y}) \\
          & = \ell_{\Omega,\bs{A}}(\pr, \bs{y}).
    \end{aligned}
\]
Equality is achieved when
\[
    \begin{aligned}
    \Omega_\Simplex^*(\bs{A}^\tr\pr)
    &= \pr^\tr\bs{A}\bs{y} - \Omega_\Simplex(\bs{y}) \iff \\
    \max_{\bs{y}' \in \Simplex^d} \pr^\tr\bs{A} \bs{y}' - \Omega(\bs{y}') &= 
    \pr^\tr\bs{A}\bs{y} - \Omega(\bs{y}),\\
    \end{aligned}
\]
where we used the fact that $\bs{y}\in\Simplex^d$. The second part of the claim
follows.

\paragraph{Property 2.}
To prove convexity in $\pr$, we rewrite the loss, for fixed $\bs{y}$, as 
\[
    \ell_{\Omega, \bs{A}}(\pr) = h(\bs{A}^\tr \pr) + \text{const},
    \quad\text{where}\quad
    h(\pt) =
    \Omega_\Simplex^*(\pt) - \pt^\tr\bs{y}.
\]
$\Omega_\Simplex^*$ is a convex conjugate, and thus itself convex.
Linear functions are convex, and the sum of two convex functions is convex,
therefore $h$ is convex.
Finally, the composition of a convex function with a linear function
is convex as well, thus the function
$\left(h\bs{A}^\tr\right)$ is convex.  Convexity of $\ell_{\Omega, \bs{A}}$ in
$\pr$ directly follows.
Convexity in $\bs{y}$ is straightforward, as the sum of a convex and a linear
function \citep[Sections 3.2.1, 3.2.2, 3.3.1]{boyd}.

\paragraph{Property 3.}
This follows from the scaling property of the convex conjugate \citep[Section 3.3.2]{boyd}
\[(t\Omega)^*(\bs{\theta}) = t\Omega^*(t^{-1}\bs{\theta})\]
Denoting $\pr' = t^{-1}\pr$, we have that
\[
\begin{aligned}
  \ell_{t\Omega, \bs{A}}(\pr, \bs{y})
  &= (t{\Omega_\Simplex})^*(\bs{A}^\tr\pr)+t\Omega_\Simplex(\bs{y})-\pr^\tr\bs{A}\bs{y}\\
  &=  t\Omega_\Simplex^*(\bs{A}^\tr\pr')+t\Omega_\Simplex(\bs{y})-\pr^\tr\bs{A}\bs{y}\\
  &= t\bigl(\Omega_\Simplex^*(\bs{A}^\tr\pr')+\Omega_\Simplex(\bs{y})-\pr'^\tr\bs{A}\bs{y}
  \bigr)
   = t \ell_{\Omega, \bs{A}}(t^{-1}\pr, \bs{y}).
\end{aligned}
\]

\end{document}